\documentclass[nohyperref]{article}

\usepackage{microtype}
\usepackage{graphicx}
\usepackage{subfigure}
\usepackage{booktabs} 

\usepackage{hyperref}

\usepackage[accepted]{icml2022}

\usepackage{amsmath}
\usepackage{amssymb}
\usepackage{mathtools}
\usepackage{amsthm}

\usepackage{url}
\usepackage{xcolor}
\usepackage{graphicx}

\usepackage{booktabs}       
\usepackage{amsfonts}       
\usepackage{nicefrac}       
\usepackage{microtype}      
\usepackage{hyperref}
\usepackage{multirow, makecell}
\usepackage[font=small,labelfont=bf]{caption} 

\usepackage[textwidth=20mm]{todonotes}
\usepackage{enumitem}

\usepackage[utf8]{inputenc}
\usepackage[russian,english]{babel}

\usepackage[capitalize,noabbrev]{cleveref}

\newcommand{\wvbert}{w2v-bert-51 (0.6B)}
\newcommand{\xlsrp}{XLS-R}
\newcommand{\xlsrpb}[1]{\xlsrp{} {(#1B)}}
\newcommand{\mslam}{mSLAM}
\newcommand{\mtlm}{mSLAM-TLM}
\newcommand{\mtlmb}[1]{\mtlm{} {(#1B)}}
\newcommand{\mctc}{mSLAM-CTC}
\newcommand{\mctcb}[1]{\mctc{} {(#1B)}}

\newcommand{\xenglish}{X $\rightarrow$ English}

\newcommand{\insertcovostshort}{
\begin{table}[hbt!]
\centering
\caption{\textbf{  Speech translation }- CoVoST 2 X$\rightarrow$En summarized results in BLEU. Full per-language results are available in the Appendix Table~\ref{tab:covost_xen_full}.}
\label{tab:covost_xen_short}
\resizebox{0.9\linewidth}{!}{ 
\begin{tabular}{l|rrrr}
\toprule
\xenglish{} & high & mid & low & all \\
\midrule 
\multicolumn{5}{l}{\it Prior work, mBART decoder init.~\citep{babu2021xls}} \\
\midrule
\xlsrpb{0.3}  & 30.6 & 18.9 & 5.1 & 13.2 \\
\xlsrpb{1}  & 34.3 & 25.5 & 11.7 & 19.3 \\
\xlsrpb{2}  & 36.1 &   27.7 &   15.1 &   22.1 \\
\midrule
\multicolumn{5}{l}{\it Our Work: Speech Only} \\
\midrule
\wvbert & 35.6 & 25.3 & 13.4 & 20.4 \\
\midrule
\multicolumn{5}{l}{\it Our Work: Speech + Text} \\
\midrule
\mtlmb{0.6} & 34.4 & 23.4 & 11.3 & 18.6 \\
\mctcb{0.6} & 35.5 & 25.2 & 13.7 & 20.6 \\
\mctcb{2} & 36.3 & 27.5 & 15.6 & 22.4 \\
\midrule
\multicolumn{5}{l}{\it Our Work: Speech Only w/ joint fine-tuning} \\
\midrule
\wvbert & 36.4 & 25.9 & 13.8 & 21.0  \\
\midrule
\multicolumn{5}{l}{\it Our Work: Speech + Text w/ joint fine-tuning} \\
\midrule
\mtlmb{0.6} & 35.5 & 25.3 & 12.3 & 19.8   \\
\mctcb{0.6} & 37.6 & 27.8 & 15.1 & 22.4  \\
\mctcb{2} & \textbf{37.8} & \textbf{29.6} & \textbf{18.5} & \textbf{24.8} \\
\bottomrule
\end{tabular} 

}
\end{table}
}

\newcommand{\insertcovost}{
\begin{table*}[hbt!]
\centering
\caption{\textbf{  Speech translation }- CoVoST 2 X$\rightarrow$En full results in BLEU.}
\label{tab:covost_xen_full}
\begin{tabular}{l|rrrr|rrrrr|rrr}
\toprule
& \multicolumn{4}{c|}{High-resource} & \multicolumn{5}{c|}{Mid-resource} & \multicolumn{3}{c}{Low-resource}\\
\midrule
\xenglish{} & fr & de & es & ca & fa & it & ru & pt & zh & tr & ar & et \\
Train Hours & 264h & 184h & 113h & 136h & 49h & 44h & 18h & 10h & 10h & 4h & 2h & 3h \\
\midrule 
\multicolumn{8}{l}{\it Prior work, mBART Decoder init.~\citep{babu2021xls}} \\
\midrule
\xlsrpb{0.3} & 32.9 & 26.7 & 34.1 & 28.7 & 5.9 & 29.0 & 26.4 & 28.3 & 4.9 & 4.6 & 3.0 & 3.5 \\
\xlsrpb{1} & 36.2 & 31.2 & 37.9 & 31.9 & 9.6  & 33.1 & 37.0 & 39.3 & 8.7 & 12.8 & 12.2 & 8.3 \\
\xlsrpb{2} &   37.6 &   33.6 &   39.2 &   33.8 &   12.9 &   34.9 & 39.5 &   41.8 &   9.4 &   16.7 &   17.1 & 11.1 \\
\midrule
\multicolumn{8}{l}{\it Our Work: Speech Only} \\
\midrule
\wvbert & 36.9 & 33.1  & 38.9 & 33.5 & 5.8 & 34.9 & 41.8 & 36.1 & 8.0 & 8.8 & 13.7 & 17.4  \\
\midrule
\multicolumn{8}{l}{\it Our Work: Speech + Text} \\
\midrule
\mtlmb{0.6} & 35.7 & 31.6 & 37.8 & 32.4 & 5.5 & 33.6 & 39.9 & 29.4 & 8.7 &  8.2 &  9.0 & 14.9  \\
\mctcb{0.6} & 36.7 & 32.7 & 39.1 & 33.4 & 6.2 & 35.0 & 41.7 & 34.2 & 8.7 & 11.7 & 13.3 & 17.2  \\
\mctcb{2}   & 37.6 & 33.8 & 39.5 & 34.4	& 8.8 & 36.1 & 43.6	& 42.0 & 7.1 & 19.7	& 15.8 & 18.6	\\
\midrule
\multicolumn{8}{l}{\it Our Work: Speech Only w/ joint fine-tuning} \\
\midrule
\wvbert & 37.5 & 34.1 & 39.6 & 34.2 & 6.1 & 35.7 & 44.1 & 34.7 & 9.0 & 12.7 & 15.5 & 19.1 \\
\midrule
\multicolumn{8}{l}{\it Our Work: Speech + Text w/ joint fine-tuning} \\
\midrule
\mtlmb{0.6} & 36.8 & 32.8 & 38.8 & 33.6 & 9.7 & 34.6 & 41.2 & 32.1 &  8.8 & 12.2 & 12.6 & 16.6 \\
\mctcb{0.6} & 38.6 & 36.1 & 40.6 & 35.2 & 7.2 & 37.0 & 47.5 & 36.4 & 10.8 & 15.6 & 14.2 & 20.3 \\
\mctcb{2}   & 39.0 & 35.9 & 41.0 & 35.4	& 9.7 & 37.3 & 48.4	& 42.8 & 10.0 & 24.2 & 19.3	& 22.6	\\
\bottomrule
\end{tabular}    
\begin{tabular}{l|rrrrrrrrr|rrrr}
\toprule
 & \multicolumn{9}{c|}{Low-resource} & \multicolumn{4}{c}{Average}\\
\midrule
\xenglish{} &  mn & nl & sv & lv & sl & ta & ja & id & cy & high & mid & low & all \\
Train Hours & 3h & 7h & 2h & 2h & 2h & 2h & 2h & 2h & 2h	& \\
\midrule 
\multicolumn{8}{l}{\it Prior work~\citep{babu2021xls}} \\
\midrule
\xlsrpb{0.3}  & 0.4 & 22.0 & 10.3 & 6.0 & 6.6 & 0.2 & 0.6 & 1.4 & 2.5 & 30.6 & 18.9 & 5.1 & 13.2 \\
\xlsrpb{1}  & 0.8 & 28.2 & 24.7 & 16.0 & 16.7 & 0.3 & 1.9 & 10.3 &  8.6 & 34.3 & 25.5 & 11.7 & 19.3 \\
\xlsrpb{2}  &   1.6 &   31.7 &   29.6 &   19.5 &   19.6 & 0.5 &   3.5 &   16.5 &   14.0 &   36.1 &   27.7 &   15.1 &   22.1 \\
\midrule
\multicolumn{8}{l}{\it Our Work: Speech Only} \\
\midrule
\wvbert & 0.3 & 33.8 & 33.9 & 16.0 & 25.5 & 0.3 & 0.9 & 3.5 & 6.2 & 35.6 & 25.3 & 13.4 & 20.4 \\
\midrule
\multicolumn{8}{l}{\it Our Work: Speech + Text} \\
\midrule
\mtlmb{0.6} & 0.5 & 31.7 & 29.5 & 14.0 & 17.4 & 0.3 & 1.7 & 3.8 & 5.1 & 34.4 & 23.4 & 11.3 & 18.6 \\
\mctcb{0.6} & 0.5 & 32.5 & 32.1 & 18.6 & 25.0 & 0.3 & 1.7 & 3.7 & 6.8 & 35.5 & 25.2 & 13.7 & 20.6 \\
\mctcb{2}   & 0.3 & 34.4 & 35.5	& 22.8 & 29.2 & 0.3	& 1.7 & 4.7	& 4.4 & 36.3 & 27.5	& 15.6 & 22.4  \\
\midrule
\multicolumn{8}{l}{\it Our Work: Speech Only w/ joint fine-tuning} \\
\midrule
\wvbert & 0.7 & 34.6 & 31.6 & 13.8 & 23.9 & 0.2 & 1.3 & 4.5 & 7.3 & 36.4 & 25.9 & 13.8 & 21.0  \\
\midrule
\multicolumn{8}{l}{\it Our Work: Speech + Text w/ joint fine-tuning} \\
\midrule
\mtlmb{0.6} & 0.3 & 33.2 & 26.3 & 15.2 & 19.8 & 0.5 & 1.3 & 3.7 & 5.6 & 35.5 & 25.3 & 12.3 & 19.8 \\
\mctcb{0.6} & 0.9 & 36.3 & 31.7 & 19.8 & 25.6 & 0.5 & 2.4 & 6.1 & 7.7 & 37.6 & 27.8 & 15.1 & 22.4  \\
\mctcb{2} &   0.8 & 37.6 & 38.5	& 26.8 & 32.3 & 0.6	& 3.3 & 8.8	& 6.7 & 37.8 & 29.6	& 18.5 & 24.8 \\
\bottomrule
\end{tabular} 
\end{table*}
}

\newcommand{\insertasr}{
\begin{table}[hbt!]
\centering
\caption{ \textbf{Speech Recognition} - Average Word Error Rate (WER) on the VoxPopuli, Babel and MLS-10Hr datasets. Per-language results can be found in Appendix Tables~\ref{tab:vp_asr}, ~\ref{tab:babel} and ~\ref{tab:mls} respectively.}
\label{tab:asr}
\resizebox{0.9\linewidth}{!}{ 
\begin{tabular}{l|rrrr}
\toprule
Model & VoxPop & Babel & MLS \\
\midrule 
\multicolumn{3}{l}{\it Prior work~\citep{babu2021xls}} \\
\midrule
\xlsrpb{0.3} & 12.8 & 32.0 & 12.8 \\
\xlsrpb{1}  & 10.6 & \textbf{29.5} & 10.9 \\
\xlsrpb{2}  & - & \textbf{29.5} & 11.0 \\
\midrule
\multicolumn{3}{l}{\it Our work: Speech-only} \\
\midrule
\wvbert & 9.3 & 32.8 & 9.9 \\
\midrule
\multicolumn{3}{l}{\it Our work: Speech + Text} \\
\midrule
\mtlmb{0.6} & 9.4 & 33.2 & 10.4 \\
\mctcb{0.6} & 9.2 & 32.9 & 10.1 \\
\mctcb{2} & \textbf{9.1} & 31.3 & \textbf{9.7} \\
\bottomrule
\end{tabular}    
}
\end{table}
}

\newcommand{\insertvp}{
\begin{table*}[bht!]
\centering
\caption{\textbf{  Speech recognition } - VoxPopuli ASR results in terms of WER.}
\begin{tabular}{l|rrrrrrrr}
\toprule
& en & de & it & fr & es & pl & ro & hu  \\
\midrule
Labeled data & 543h & 282h & 91h & 211h & 166h & 111h & 89h & 63h \\
\midrule 
\multicolumn{8}{l}{\it Prior work \citep{babu2021xls}} \\
\midrule
\xlsrpb{0.3} & 10.2 & 13.0 & 19.2 & 12.6 & 9.8 & 9.6 & 7.9 & 11.6 \\
\xlsrpb{1}  &    8.8 &   11.5 &   15.1 &   10.8 &   8.2 &   7.7 &   7.3 &   9.6 \\
\midrule
\multicolumn{8}{l}{\it Our work: Speech-only} \\
\midrule
\wvbert & 7.2 & 9.0 & 15.8 & 9.2 & 8.6 & 6.5 & 7.6 & 8.4 \\
\midrule
\multicolumn{8}{l}{\it Our work: Speech + Text} \\
\midrule
\mtlmb{0.6} & 7.3 & 8.9 & 15.6 & 9.3 & 8.7 & 6.5 & 8.5 & 8.4 \\
\mctcb{0.6} & 7.1 & 8.9 & 15.6 & 9.3 & 8.6 & 6.5 & 8.5 & 8.1  \\
\mctcb{2} & 7.0	  & 8.7 & 15.4 & 9.4 & 8.4 & 6.4 & 7.8 & 8.4 \\
\bottomrule
\toprule
& nl & cs & sl & fi & hr & sk & Avg \\
\midrule
Labeled data & 53h & 62h & 10h & 27h & 43h & 35h & \\
\midrule 
\multicolumn{8}{l}{\it Prior work \citep{babu2021xls}} \\
\midrule
\xlsrpb{0.3} & 14.8 & 10.5 & 24.5 & 14.2 & 12.3 & 8.9 & 12.8\\
\xlsrpb{1}  &   12.5 &   8.7 &   19.5 &   11.3 &   10.0 &   7.1 &   10.6 \\
\midrule
\multicolumn{8}{l}{\it Our work: Speech-only} \\
\midrule
\wvbert & 10.5 & 7.0	& 15.8 & 9.3 & 9.1	& 6.0	& 9.3\\
\midrule
\multicolumn{8}{l}{\it Our work: Speech + Text} \\
\midrule
\mtlmb{0.6} & 10.5 & 7.1 & 15.8 & 9.0 & 10.0 & 6.2 & 9.4\\
\mctcb{0.6} & 10.3 & 7.0 & 14.2 & 9.2 & 9.1 & 5.9 & 9.2\\
\mctcb{2} &10.5	& 6.8 & 15.1 & 8.7 & 9.1 & 6.0 & \textbf{9.1}\\
\bottomrule
\end{tabular}    
\label{tab:vp_asr}
\end{table*}
}

\newcommand{\insertbabel}{
 \begin{table*}[bht!]
        \begin{center}
            \captionof{table}{\textbf{  Speech recognition }- BABEL ASR baselines in five languages, reporting WER.
            \label{tab:babel}}
            \resizebox{0.65\linewidth}{!}{
            \begin{tabular}[t]{l|ccccc|c}
            \toprule
            {  Model} & {  as} & {  tl} & {  sw} & {  lo} & {  ka} & {  Avg} \\
                \midrule
                \midrule
                \multicolumn{1}{l|}{Number of pretraining hours} & 55h & 76h & 30h & 59h  & 46h & - \\
                \multicolumn{1}{l|}{Number of fine-tuning hours} & 55h & 76h & 30h & 59h & 46h & - \\
                \midrule
                \midrule
                \multicolumn{7}{l}{\it Prior work (with LM)~\citep{babu2021xls}} \\
                \midrule
                \xlsrpb{0.3} & 42.9 & 33.2 & 24.3 & 31.7 & 28.0 & 32.0 \\
                \xlsrpb{1} & 40.4 & 30.6 & 21.2 & 30.1 & 25.1 & 29.5 \\
                \xlsrpb{2} & \textbf{39.0} & \textbf{29.3} & \textbf{21.0} & \textbf{29.7} & \textbf{24.3} & \textbf{28.7} \\
                \midrule
                \multicolumn{7}{l}{\it Our work: Speech-only, no LM} \\
                \midrule
                \wvbert & 42.8 & 32.9 & 26.7 & 30.6 & 31.1 & 32.8 \\
                \midrule
                \multicolumn{7}{l}{\it Our work: Speech + Text, no LM} \\
                \midrule
                \mtlmb{0.6} & 43.0 & 32.7 & 27.6 & 30.9 & 31.8 & 33.2 \\
                \mctcb{0.6} & 42.7 & 32.6 & 27.1 & 30.7 & 31.4 & 32.9 \\
                \mctcb{2} & 41.1 & 31.1 & 25.1 & 29.9 & 29.1 & 31.2 \\
                \bottomrule
            \end{tabular}
            }
        \end{center}
    \end{table*}
}

\newcommand{\insertmls}{
\begin{table*}[bht!]
        \begin{center}
            \caption{\textbf{  Speech recognition }- Multilingual LibriSpeech (MLS) ASR baselines in 8 languages, reporting WER. \label{tab:mls}}
            \resizebox{0.8\linewidth}{!}{
            \begin{tabular}[h!]{l|cccccccc|c}
            \toprule
                {  Model} & {  en }& {  de }& {  nl }& {  fr }& {  es }& {  it }& {  pt} & {  pl} & {  Avg} \\
                \midrule
                \midrule
                \multicolumn{1}{l|}{Number of training hours} & 10 & 10 & 10 & 10 & 10 & 10 & 10 & 10 & -  \\
                \midrule
                \multicolumn{9}{l}{\it Prior work (monolingual fine-tuning)~\citep{babu2021xls}} \\
                \midrule
                XLS-R(0.3B) & 15.9 & 9.0 & 13.5 & 12.4 & 8.1 & 13.1 & 17.0 & 13.9 & 12.8 \\
                XLS-R(1B) & 12.9 & 7.4 & \textbf{11.6} & 10.2 & 7.1 & 12.0 & 15.8 & 10.5 & 10.9 \\
                XLS-R(2B) & 14.0 & 7.6 & \textbf{11.8} & 10.0 & 6.9 & 12.1 & 15.6 & 9.8 & 11.0 \\
                \midrule
                \multicolumn{9}{l}{\it Our work: Speech Only (multilingual fine-tuning)} \\
                \midrule
                \wvbert & 12.7 & 7.0 & 12.6 & 8.9 & 5.9 & 10.3 & 14.6 & \textbf{6.9} & 9.9 \\
                \midrule
                \multicolumn{9}{l}{\it Our work: Speech + Text (multilingual fine-tuning)} \\
                \midrule
                \mtlmb{0.6} & 13.9 & 7.2 & 13.0 & 9.9 & 5.8 & 10.7 & \textbf{14.2} & 8.4 & 10.4 \\
                \mctcb{0.6} & 13.3 & 7.0 & 12.5 & 9.7 & \textbf{5.5} & 10.5 & \textbf{14.1} & 8.5 & 10.1 \\
                \mctcb{2} & \textbf{11.9} & \textbf{6.6} & 12.4 & \textbf{8.5} & 5.8 & \textbf{9.8} & 15.2 & 7.7 & \textbf{9.7} \\
                \bottomrule
            \end{tabular}
            }
       \vspace{-0.4cm}
        \end{center}
    \end{table*}
}

\newcommand{\insertsc}{
\begin{table}[bht!]
        \begin{center}
            \caption{{\bf Speech Classification }- MINDS-14 speech intent classification and Fleurs speech language identification accuracy. \label{tab:sc}}
            \resizebox{0.9\linewidth}{!}{
            \begin{tabular}[h!]{l|c|c}
            \toprule
                {\bf Model} & {\bf MINDS-14}& {\bf Fleurs-LangID} \\
                \midrule
                \midrule
                \multicolumn{3}{l}{\it Our work: Speech Only} \\
                \midrule
                \wvbert & 82.7 & 71.4 \\
                \midrule
                \multicolumn{3}{l}{\it Our work: Speech + Text} \\
                \midrule
                \mtlmb{0.6} & 84.0 & 76.0 \\
                \mctcb{0.6} & \textbf{86.9} & 73.3 \\
                \mctcb{2} & {86.6} & \textbf{77.7} \\
                \bottomrule
            \end{tabular}
            }
      \vspace{-0.4cm}
        \end{center}
    \end{table}
}

\newcommand{\inserttcshort}{
\begin{table}[bht!]
        \begin{center}
            \caption{{\bf Text Classification }- XNLI dev accuracy on English, European (bg, de, el, es, fr) and Non-European (ar, hi, ru, sw, th, tr, ur, vi, zh) languages. Full results in Appendix Table~\ref{tab:xnli}. Note, for \mslam{} models, only $450M$ and $1.4B$ out of the $600M$ and $2B$ parameters are fine-tuned for text tasks. \label{tab:xnli_short}}
            \resizebox{0.95\linewidth}{!}{
            \begin{tabular}[h!]{lcccc}
            \toprule
                {\bf Model} & \bf{English} &	\bf{Euro} &	\bf{Non-Euro} &	\bf{Avg} \\
                \midrule
                \midrule
                \multicolumn{5}{l}{\it Prior work: Text Only, Zero-shot~\cite{xue-etal-2021-mt5}} \\
                \midrule
                mT5-Small (0.3B) & 79.6 & 66.6 & 60.4 & 63.8\\
                mT5-Base (0.6B) & 84.5 & 77.1 & 69.5 & 73.0\\
                \midrule
                \multicolumn{5}{l}{\it Our work: Speech + Text, Zero-shot} \\
                \midrule
                \mtlmb{0.6} & 75.7 & 57.5 &	48.6 & 53.4\\
                \mctcb{0.6}& 80.4 & 71.4 &	49.5 & 58.9\\
                \mctcb{2} & 80.1 &74.4 & 59.9 &	66.1 \\
                \midrule
                \midrule
                \multicolumn{5}{l}{\it Prior work: Text Only, Translate-Train-All~\cite{xue-etal-2021-mt5}} \\
                \midrule
                mT5-Small (0.3B) & 78.3 & 73.6 & 69.2 & 71.3\\
                mT5-Base (0.6B) & 85.9 & 82.1 & 77.9 & 79.8\\
                \midrule
                \multicolumn{5}{l}{\it Our work: Speech + Text, Translate-Train-All} \\
                \midrule
                \mtlmb{0.6} & 74.1 & 69.3 &	64.6 & 66.8 \\
                \mctcb{0.6} & 81.1 & 76.0 & 	65.5 & 70.0 \\
                \mctcb{2} & 84.1 & 80.5	& 73.7 & 76.1\\
                \bottomrule
            \end{tabular}
            }
       \vspace{-0.4cm}
        \end{center}
    \end{table}
}

\newcommand{\inserttc}{
\begin{table*}[bht!]
        \begin{center}
            \caption{{\bf Text Classification }- XNLI dev accuracy for all 15 languages. For \mslam{} models, only $450M$ and $1.4B$ out of the $600M$ and $2B$ parameters are fine-tuned for XNLI. \label{tab:xnli}}
            \resizebox{0.95\linewidth}{!}{
            \begin{tabular}[h!]{l|c|cccccccccccccc|c}
            \toprule
                {\bf Model} & \bf{en} &	\bf{ar} &	\bf{bg} &	\bf{de} &	\bf{el} &	\bf{es} &	\bf{fr} &	\bf{hi} &	\bf{ru} &	\bf{sw} &	\bf{th} &	\bf{tr} &	\bf{ur} &	\bf{vi} &	\bf{zh} &	\bf{Avg} \\
                \midrule
                \midrule
                \multicolumn{9}{l}{\it Prior work: Text Only, Zero-shot~\cite{xue-etal-2021-mt5}} \\
                \midrule
                mT5-Small (0.3B) & 79.6 & 62.2 & 67.8 &	64.8 & 65.8 & 68.4 & 66.2 & 59.0 & 65.3 &	55.4 & 63.2 & 58.9 & 54.5 & 61.8 & 63.4 & 63.8\\
                mT5-Base (0.6B) & 84.5 & 71.2 & 76.9 &	75.6 & 76.3 & 79.0 & 77.7 & 66.9 & 74.9 &	63.6 & 70.0 & 69.2 & 64.8 & 72.0 & 72.5 & 73.0\\
                \midrule
                \multicolumn{9}{l}{\it Our work: Speech + Text, Zero-shot} \\
                \midrule
                \mtlmb{0.6} & 75.7&47.3&56.7&	55.1	& 52.2&60.9&62.8&48.6&	58.5& 	46.0&	46.9&	51.3&	47.2&	50.7&41.0 & 53.4\\
                \mctcb{0.6}& 80.4 & 46.5 & 69.8&72.1&	67.5&74.7 & 72.9 & 42.0 & 68.7 & 45.5 &	42.9 & 48.7 & 44.2 & 63.3 &	43.3  & 58.9\\
                \mctcb{2} & 80.1 & 61.1 & 73.3 & 74.7 & 72.7 & 76.0 & 75.3 & 59.4 & 70.9 & 52.2 & 56.8 & 63.9 & 59.0 & 65.9 & 50.1 & 66.1\\
                \midrule
                \midrule
                \multicolumn{9}{l}{\it Prior work: Text Only, Translate-Train-All~\cite{xue-etal-2021-mt5}} \\
                \midrule
                mT5-Small (0.3B) & 78.3 & 68.8 & 73.5 &	73.2 & 73.4 & 74.4 & 73.5 & 67.4 & 71.1 &	67.2 & 71.1 & 69.9 & 63.6 & 70.5 & 72.9 & 71.3\\
                mT5-Base (0.6B) & 85.9 & 78.8 & 82.2 &	81.6 & 81.4 & 83.0 & 82.1 & 77.0 & 81.1 &	74.8 & 78.6 & 78.4 & 73.3 & 78.9 & 80.2 & 79.8\\
                \midrule
                \multicolumn{9}{l}{\it Our work: Speech + Text, Translate-Train-All} \\
                \midrule
                \mtlmb{0.6} & 74.3& 64.2 & 68.7 & 69.5 & 69.2 &	70.2 &	71.4 &	64.5 &	65.4 &	63.4 &	65.6 &	65.9 &	62.4 &	67.3 &	64.4 & 67.1 \\
                \mctcb{0.6} & 81.1&63.5&76.7&76.0&73.1&77.8&76.4&63.6&73.1&64.1&64.9&66.8&60.5&68.4&64.5 & 70.0 \\
                \mctcb{2} & 84.1 &	80.2 &	80.1 &	78.7 &	82.9 &	80.5 &	74.4 &	72.1 &	76.8 &	71.7 &	73.8 &	76.2 &	69.8 &	75.9 &	72.8 & 76.1 \\
                \bottomrule
            \end{tabular}
            }
       \vspace{-0.4cm}
        \end{center}
    \end{table*}
}
\newcommand{\insertanalysis}{
\begin{table}[bht!]
\caption{{\bf Zero-shot Performance }- CoVoST 2 translation results with X$\rightarrow$Y indicating X as the fine tuning modality and Y as the testing modality: S=Speech, T=Text. CAE is our CTC zero-shot character auto-encoding probe.} \label{tab:crossmodalanalysis}
\centering
\resizebox{0.8\linewidth}{!}{
\begin{tabular}{cr|rrr|r}
 \toprule
     & \multicolumn{1}{c|}{Hours}  & \multicolumn{3}{c|}{BLEU $\uparrow$}                                                                                        & \multicolumn{1}{c}{CER $\downarrow$}                 \\
Lang & \multicolumn{1}{l|}{Paired} & \multicolumn{1}{l}{S$\rightarrow$S} & \multicolumn{1}{l}{S$\rightarrow$T} & \multicolumn{1}{l|}{T$\rightarrow$S} & \multicolumn{1}{l}{S$\rightarrow$T CAE} \\
\midrule
ar   & 0                               & 13.3                    & 0.0                     & 0.0                     & 82.6                        \\
fa   & 0                               & 6.2                     & 0.0                     & 0.0                     & 80.0                        \\
ja   & 0                               & 1.6                     & 0.0                     & 0.0                     & 100.0                       \\
zh   & 0                               & 8.7                     & 0.0                     & 0.0                     & 100.0                       \\
cy   & 0                               & 6.1                     & 0.1                     & 0.0                     & 24.3                        \\
mn   & 0                               & 0.5                     & 0.1                     & 0.0                     & 78.4                        \\
id   & 0                               & 3.9                     & 5.1                     & 0.0                     & 10.4                        \\
lv   & 0                               & 19.4                    & 8.2                     & 0.0                     & 18.4                        \\
et   & 0                               & 17.2                    & 8.3                     & 0.0                     & 16.5                        \\
sv   & 0                               & 33.1                    & 15.2                    & 0.0                     & 13.9                        \\
ca   & 0                               & 33.4                    & 16.7                    & 0.0                     & 10.0                        \\
ru   & 0                               & 41.7                    & 21.9                    & 0.0                     & 85.9                        \\
sl   & 6                               & 24.9                    & 7.8                     & 0.0                     & 10.6                        \\
pt   & 10                              & 34.2                    & 17.2                    & 0.0                     & 9.0                         \\
nl   & 41                              & 32.6                    & 16.8                    & 0.0                     & 11.3                        \\
ta   & 63                              & 0.3                     & 0.0                     & 0.0                     & 91.2                        \\
tr   & 69                              & 11.7                    & 1.7                     & 0.0                     & 12.6                        \\
it   & 79                              & 35.0                    & 19.7                    & 0.0                     & 11.2                        \\
es   & 140                             & 39.1                    & 21.2                    & 0.0                     & 7.9                         \\
fr   & 179                             & 36.7                    & 20.0                    & 0.0                     & 9.4                         \\
de   & 197                             & 32.7                    & 16.8                    & 0.0                     & 8.3                        \\
\bottomrule
\end{tabular}
}
\end{table}
}

\newcommand{\insertCtcExamples}{
\begin{table*}[bth!]
\centering
\caption{{\bf CTC Probing Examples }- CoVoST CTC Probe with zero-shot text input to visualize zero-shot text encodings. Gold is the desired output as well as the text input. Romanization is provided by the GOST 7.79 System B standard for Cyrillic transliteration.} \label{tab:analysisExamples}
\resizebox{0.8\linewidth}{!}{
\begin{tabular}{cl|l}
 \toprule
\multirow{2}{*}{fr} & Gold & \texttt{Certains départements sont mieux équipés que d’autres.} \\
 & S$\rightarrow$T (CAE) & \texttt{certains départements sont mieux équipés que d'autres} \\
 \midrule
\multirow{3}{*}{ru} & Gold & \foreignlanguage{russian}{И нам следует руководствоваться им.ль.} \\
 & Romanized Gold & \texttt{I nam sleduet rukovodstvovat'sya im.} \\
 & S$\rightarrow$T (CAE) & \texttt{ {} nam sleduet rucovodstvowats {} {} {} im.} \\
\bottomrule
\end{tabular}
}
\end{table*}
}

\newcommand{\insertablate}{
\begin{table}[hbt!]
\centering
\caption{Comparing \mctc models trained with and without unlabeled text on CoVoST-2 ST BLEU (with joint fine-tuning), MINDS-14 accuracy and Fleurs-LangID accuracy.}
\label{tab:ablate}
\resizebox{0.9\linewidth}{!}{ 
\begin{tabular}{l|ccc}
\toprule
& CoVoST Avg. & MINDS-14 & Fleurs \\
\midrule 
\wvbert & 21.0 & 82.7 & 71.4 \\
\midrule 
\mctcb{0.6} & 22.4 & 86.9 & 73.3\\
\midrule
\mctcb{0.6} - Text & 22.2 & 85.0 & 71.9  \\
\bottomrule
\end{tabular} 

}
\end{table}
}

\newcommand{\insertTranslationExamples}{
\begin{table}[bth!]
\centering
\caption{{\bf Zero-shot text translation examples.} Drawn from the CoVoST 2 FrEn test set, decoded by \mctcb{0.6}.} \label{tab:translationExamples}
\resizebox{1.0\linewidth}{!}{
\begin{tabular}{l|l}
 \toprule
Source & {Il réalise aussi quelques courts-métrages.} \\
Gold   & {He also makes short films.} \\
S$\rightarrow$T & {He also writes a few short films either short films either short films.} \\
 \midrule
Source & {Il a réalisé deux courts-métrages.} \\
Gold   & {He produced two short films.} \\
S$\rightarrow$T & {He created two short short films.} \\
\bottomrule
\end{tabular}
}
\end{table}
}

\icmltitlerunning{mSLAM: Massively multilingual joint pre-training for speech and text}

\begin{document}

\twocolumn[
\icmltitle{\mslam: Massively multilingual joint pre-training for speech and text }



\icmlsetsymbol{equal}{*}

\begin{icmlauthorlist}
\icmlauthor{Ankur Bapna}{equal,goog}
\icmlauthor{Colin Cherry}{equal,goog}
\icmlauthor{Yu Zhang}{equal,goog}
\icmlauthor{Ye Jia}{goog}
\icmlauthor{Melvin Johnson}{goog}
\icmlauthor{Yong Cheng}{goog}
\icmlauthor{Simran Khanuja}{goog}
\icmlauthor{Jason Riesa}{goog}
\icmlauthor{Alexis Conneau}{goog}
\end{icmlauthorlist}

\icmlaffiliation{goog}{Google, USA}

\icmlcorrespondingauthor{Ankur Bapna}{ankurbpn@google.com}
\icmlcorrespondingauthor{Colin Cherry}{colincherry@google.com}
\icmlcorrespondingauthor{Yu Zhang}{ngyuzh@google.com}

\icmlkeywords{Machine Learning, ICML}

\vskip 0.3in
]



\printAffiliationsAndNotice{\icmlEqualContribution} 

\begin{abstract}
We present \mslam{}, a multilingual Speech and LAnguage Model that learns cross-lingual cross-modal representations of speech and text by pre-training jointly on large amounts of unlabeled speech and text in multiple languages. \mslam{} combines w2v-BERT pre-training on speech with SpanBERT pre-training on character-level text, along with Connectionist Temporal Classification (CTC) losses on paired speech and transcript data, to learn a single model capable of learning from and representing both speech and text signals in a shared representation space. We evaluate \mslam{} on several downstream speech understanding tasks and find that joint pre-training with text improves quality on speech translation, speech intent classification and speech language-ID while being competitive on multilingual ASR, when compared against speech-only pre-training. Our speech translation model demonstrates zero-shot text translation without seeing any text translation data, providing evidence for cross-modal alignment of representations. \mslam{} also benefits from multi-modal fine-tuning, further improving the quality of speech translation by directly leveraging text translation data during the fine-tuning process. Our empirical analysis highlights several opportunities and challenges arising from large-scale multimodal pre-training, suggesting directions for future research.
\end{abstract}

\section{Introduction}
\label{sec:introduction}
Multilingual pre-trained models have demonstrated large quality gains on a variety of multilingual Natural Language Processing (NLP) tasks~\citep{hu2020xtreme,ruder2021xtreme}. With the emergence of multilingual pre-trained models of speech like XLSR~\citep{conneau2020unsupervised,babu2021xls}, similar improvements have also been observed on speech understanding tasks. One key advantage of multilingual pre-trained models is the ability to overcome data skew across languages to improve quality on low resource languages~\citep{arivazhagan2019massively}. By training a shared set of (usually attention-based) parameters on many languages, these models can learn crosslingually aligned representations of text or speech in a shared representation space~\citep{kudugunta2019investigating,wu2019emerging}. These shared multilingual representations allow multilingual pre-trained models to use supervised data in one language to benefit lower-resource languages~\citep{conneau2019cross}. An extreme scenario of cross-lingual transfer learning is zero-shot transfer, where using supervised data to fine-tune a pre-trained model on a source language exhibits non-zero performance on a target language; without utilizing any supervision for the target language
~\cite{johnson2017google,conneau2018xnli}.

Given the convergence of architectures~\citep{vaswani2017attention} and objectives~\citep{devlin2019bert,baevski2020wav2vec,chung2021w2v} across the speech and text modalities, building a single model that could learn cross-lingual cross-modal representations of speech and text from hundreds of languages is the next natural step. Such a model can enable transfer learning across the two modalities, directly benefiting languages (and domains) with limited amounts of speech or text data. In addition, joint models of speech and text can likely enable end-to-end speech understanding tasks directly from the speech signal, including tasks like speech translation, speaker intent classification and speech language-identification, bypassing errors introduced by an intermediate automatic speech recognition (ASR) system.

While there are several potential advantages from multilingual pre-trained models of speech and text, these models suffer from interference and capacity dilution~\citep{bapna2021slam}. This effect has also been documented in multilingual pre-trained models of text. While lower resource languages benefit from transfer learning, with increasing multilinguality, high resource languages lose quality~\citep{Caruana1997,arivazhagan2019massively,conneau2019unsupervised}. This deterioration is typically addressed by either increasing model capacity~\citep{arivazhagan2019massively,babu2021xls} or incorporating approaches from multi-task learning to reduce interference by leveraging architectural or optimization improvements~\cite{wang2020gradient,raffel2019exploring}. 

In this work we present \mslam{}, a multilingual pre-trained model of speech and text that has been pre-trained with speech from $51$ languages and text from $101$ languages. \mslam{} is a multilingual extension of SLAM~\citep{bapna2021slam}, with the addition of a Connectionist Temporal Classification (CTC) loss~\citep{graves2006connectionist} on the paired speech-text data, to reduce interference and encourage stronger alignment across the two modalities.

On several downstream speech understanding tasks, including CoVoST-2 21$\rightarrow$En speech translation~\cite{wang2020covost},  Fleurs speech language identification (Section~\ref{subsec:tasks-sc}) and Minds-14 speech intent classification~\cite{gerz2021multilingual}, \mslam{} demonstrates significant quality improvements over equivalent models trained only on speech. On multilingual ASR tasks, including MLS-10Hr~\citep{pratap2020mls}, VoxPopuli~\citep{wang2021voxpopuli} and Babel~\citep{Gales2014SpeechRA}, \mslam{} matches the performance of the speech-only baseline. We also evaluate \mslam{} on XNLI~\cite{conneau2018xnli}, to understand its strengths and limitations on text tasks.
We find that the addition of the CTC loss significantly improves quality on several speech and text understanding tasks, highlighting the importance of alleviating interference in multi-modal pre-trained models.

We also conduct analyses to understand the extent of multi-modal representation alignment in \mslam{}. When fine-tuned with only speech translation data, \mslam{} is capable of zero-shot text translation in several languages, suggesting that the model is capable of learning from data in one modality to improve quality in the other. \mslam{} also benefits from multi-modal supervised data. On CoVoST-2, we jointly fine-tune \mslam{} on multilingual speech translation and text translation, further improving speech translation quality by $2$ BLEU; improving over a significantly larger \xlsrpb{2} model~\citep{babu2021xls} and establishing a new state of the art on this dataset. Increasing \mslam{} model capacity to $2B$ parameters results in further quality improvements on most downstream tasks.






\section{Background}
\label{sec:background}

\textbf{Multimodal pre-training:}
SLAM~\cite{bapna2021slam} is a multimodal speech and text pretraining method, which trains a single Conformer~\cite{gulati2020conformer} with SpanBERT~\citep{Joshi2020SpanBERTIP} and w2v-BERT~\citep{chung2021w2v} self-supervised losses that leverage unlabeled monomodal data, as well as a TLM loss~\cite{conneau2019cross,zheng2021fused} and a speech-text matching loss~\citep{li2021align} that both use supervised speech recognition data. Pre-trained speech representations have been shown to be close to text~\cite{baevski2021unsupervised} and SLAM leverages this similarity for cross-modal transfer. Compared to mono-modal pre-trained models, SLAM shows improvements on speech translation, similar performance on speech recognition but degradation on text downstream tasks, exposing a transfer-interference trade-off that has been previously studied in multilingual models~\cite{arivazhagan2019massively}. Because SLAM focuses on English it is harder to notice cross-modal transfer, as both modalities have a large amount of unlabeled data. In many languages, speech data is scarcer than text, or vice-versa. In this scenario cross-modal transfer is more likely, similar to how high-resource languages transfer to low-resource languages in multilingual pre-training. \mslam{} exploits both cross-lingual and cross-modal transfer by simultaneously training on both modalities in a large number of languages.

\textbf{Multilingual pre-training:}
In multilingual understanding literature, models like mBERT~\cite{devlin2019bert}, XLM-R~\cite{conneau2019cross} or mT5~\cite{xue-etal-2021-mt5} have shown the benefit of cross-lingual transfer for improving representations of low-resource languages: on these languages, multilingual models strongly outperform monolingual pre-trained models on public benchmarks~\cite{conneau2018xnli,conneau2019unsupervised,lewis2019mlqa,hu2020xtreme,ruder2021xtreme}. Past work has also leveraged parallel data to improve multilingual text representations, e.g. with TLM~\cite{conneau2019cross}, explicit alignment~\cite{hu-etal-2021-explicit} or nmT5~\cite{kale2021nmt5}. Similarly, in speech understanding, multilingual pre-trained models ~\cite{kawakami2020learning,conneau2020unsupervised,babu2021xls} based on self-supervised losses ~\cite{oord2018representation,baevski2020wav2vec} improve representations of low-resource languages at the cost of reduced performance on high-resource languages.  In particular, multilingual pre-trained models like XLS-R expanded the few-shot learning capability of wav2vec 2.0 ~\cite{xu2021self} to many other languages, both for speech recognition and speech translation~\cite{wang2020covost}. Interestingly, for speech, no lexical overlap across languages is leveraged during training, but multilingual representations still emerge from parameter sharing of the Transformer network~\cite{wu2019emerging}. Leveraging text can potentially create connections between speech representations across languages through shared text anchor embeddings of identical character strings. Past work also leverages supervised ASR data to build multilingual representations of speech~\cite{kannan2019large,bai2021joint}, similar to how multilingual machine translation in NLP is used to build multilingual representations~\cite{eriguchi2018zeroshot,siddhant2020evaluating}.

\begin{figure*}[t]
	\begin{center}
          \includegraphics[width=\linewidth]{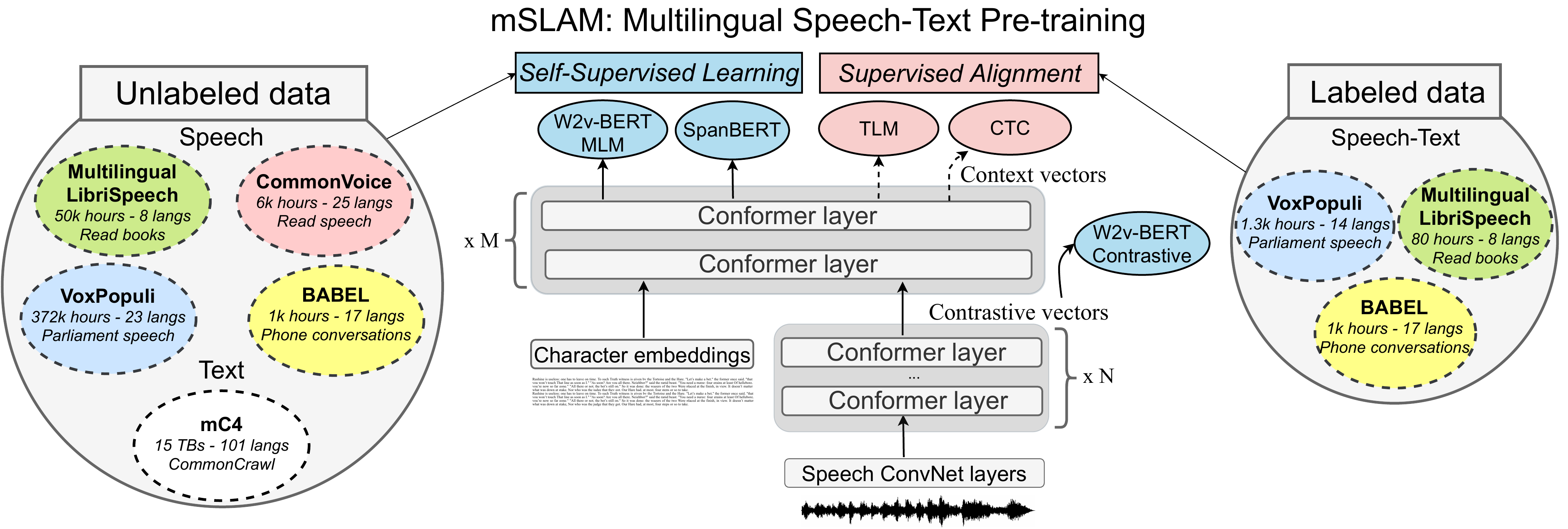}
        \captionof{figure}{\textbf{Multilingual Speech-Text Pretraining} We pre-train a large multilingual speech-text Conformer on 429K hours of unannotated speech data in 51 languages, 15TBs of unannotated text data in 101 languages, as well as 2.3k hours of speech-text ASR data.
        \label{fig:modelone}}
        \vspace{-0.2cm}
	\end{center}
\end{figure*}

\section{Pre-training Method}
\label{sec:pretraining}

\subsection{Architecture and Objectives}
\label{subsec:pretrain-arch}

Our pre-training approach builds on SLAM~\citep{bapna2021slam} and extends it to the massively multilingual setting. Specifically, we build on SLAM-TLM, that combines together pre-training on speech unlabeled data with w2v-BERT~\cite{chung2021w2v}, text with spanBERT~\cite{Joshi2020SpanBERTIP} and on paired speech-transcript data with TLM~\cite{conneau2019cross}. We skip the Speech-Text-Matching (STM) task since preliminary experiments didn't reveal any advantages over SLAM-TLM.


\mslam{} pre-training differs from SLAM on a couple of points. First, instead of using $32k$ token sentence-piece tokenization~\citep{kudo-richardson-2018-sentencepiece}, we use a character vocabulary, containing $4096$ tokens spanning $101$ languages. This results in longer sequence lengths, which we cap to $512$ characters. We also increase the length of masked spans from $5$ to $20$ tokens for the spanBERT objective. Second, we apply a CTC loss~\citep{graves2006connectionist,graves2014towards} on the speech portion of the paired input, using the character-level transcript as the target. This CTC loss is applied in addition to TLM, so the input consists of a concatenated masked speech and masked text sequence, with the CTC loss applied to the speech portion of the output. We share the softmax vocabulary and parameters used for CTC with the softmax used for training the spanBERT objective with text input. We find that this CTC loss ensures stronger alignment between the speech and text representations learnt by the model, as further discussed in Sections~\ref{sec:results} and~\ref{sec:analysis}.

In our $2B$ model, we increase the model dimension from $1024$ to $1408$, and the number of conformer layers from $24$ to $40$. We keep $8$ layers in the contrastive block, and $32$ in the MLM block. The peak learning rate is reduced from $6.0e-4$ to $3.6e-4$ for better training stability. Other hyper-parameters remain the same as the base $600M$ model. Note that our $2B$ model contains close $1.84B$ parameters.

\subsection{Pre-training Data}
\label{subsec:pretrain-data}

We use three types of data for pre-training \mslam{}; unlabeled speech drawn from multiple public datasets, unlabeled text from mC4~\citep{Xue2021mT5AM} and paired speech and text transcript data from multiple sources.

\subsubsection{Unlabeled Speech Data}
\label{subsubsec:pretrain-data-speech}
We use approximately $429k$ hours of unlabeled speech data in $51$ languages\footnote{Counting languages with more than $1$ hour of speech data.}. Our unlabeled speech data closely follows the pre-training data used for \xlsrp~\citep{babu2021xls} with one major difference: we do not use VoxLingua. As a consequence our model is pre-trained on speech from $51$ languages as compared to $128$ for \xlsrp, and our pre-training set is smaller by $6.6k$ hours.

We train on $372k$ hours of speech data spanning 23 languages from VoxPopuli~\citep{wang2021voxpopuli}, read speech data in 25 languages drawn from the v6.1 release of Common Voice~\citep{ardila2019common}, $50k$ hours of read books data in eight European languages from Multilingual LibriSpeech~\citep{pratap2020mls} and $1k$ hours of telephonic conversation data spanning $17$ African and Asian languages from BABEL~\citep{Gales2014SpeechRA}.

\subsubsection{Unlabeled Text Data}
\label{subsubsec:pretrain-data-text}
For pre-training with unlabeled text, we use the mC4 dataset~\citep{xue-etal-2021-mt5} spanning $101$ languages. We upsample lower resource languages using temperature-based sampling~\citep{arivazhagan2019massively}, with $T=3.0$.

\subsubsection{Paired Speech-Transcript Data}
\label{subsubsec:pretrain-data-paired}
In addition to training with unlabeled speech and text, we also use approximately $2.4k$ hours of paired speech and transcript data spanning $32$ languages, for training with the CTC and TLM alignment losses. This data is drawn from the following sources:

\textbf{VoxPopuli:} Approximately $1.3k$ hours of speech and transcript data spanning 14 languages. We exclude languages with less than $1$ hour of data following~\citet{wang2021voxpopuli}.

\textbf{Multilingual LibriSpeech (MLS):} We use the $10$ hour training splits of the paired data for each of the $8$ MLS languages. We exclude any paired data outside the $10$ hour splits to align with our downstream evaluations.

\textbf{Babel:} $1k$ hours of speech and transcript data spanning $17$ languages from the Babel ASR task.

\subsection{Optimization and Hyperparameters}
\label{subsec:pretrain-opt}
At each training step, we train \mslam{} on all three types of data; each batch is composed of $2048$ sequences of unlabeled speech, $8192$ sequences of text and $256$ sequences of paired data. Our speech-only baseline, \wvbert, sees a batch composed of $4096$ unlabeled speech sequences at every step. For our best run based on CoVoST dev performance, the speech loss has a coefficient of $1.0$, the text loss has a coefficient of $0.3$ and the paired CTC loss has a coefficient of $0.03$ (to avoid over-fitting to the small paired data). We use the Adam optimizer~\citep{kingma2014adam} with a Transformer learning rate schedule~\citep{vaswani2017attention}. We use $40k$ warmup steps, linearly increasing the learning rate to $6\times10^{-4}$, followed by inverse square root decay. We train all $600M$ models for $1.3m$ steps. The $2B$ model was pre-trained for $350k$ steps.

\section{Downstream tasks}
\label{sec:tasks}

\subsection{Multilingual Speech Translation}
\label{subsec:tasks-ast}
\textbf{CoVoST 2 Speech Translation:} CoVoST 2 \cite{wang2020covost} is a multilingual speech translation (ST) dataset created by professional translation of the Common Voice speech corpus \citep{ardila2020common}. The audio consists of read speech crowd-sourced through the Mozilla Common Voice project. We evaluate on a multilingual XX-to-English task that covers translation from 21 source languages into English. The training data ranges in size from 264 hours speech for French to about 1 hour speech for Indonesian. 

\textbf{Multi-modal fine-tuning:} Apart from fine-tuning with just ST data, we leverage the ability of \mslam{} to learn from both speech and text modalities by using text translation data in addition to the CoVoST 2 ST data for multi-modal joint fine-tuning. For each CoVoSt 2 XX-to-English language pair, we use the text translation data from CoVoST 2 combined with all data from either WMT or TED Talks, as available. Specifically, we pair with WMT20~\cite{barrault-etal-2020-findings} for ja, ta,  WMT19~\cite{barrault-etal-2019-findings} for de, ru, zh, WMT18~\cite{bojar-etal-2018-findings} for et, tr,  WMT17~\cite{bojar-etal-2017-findings} for lv, WMT15~\cite{bojar-etal-2015-findings} for fr, WMT13~\cite{bojar-etal-2013-findings} and TED59~\cite{qi-etal-2018-pre} for ar, fa, id, it, nl, pt, sl, sv, leaving ca and cy unpaired. 
%
%

%
We attach a 6-layer, 512-dimension Transformer decoder to our pre-trained encoders. This decoder has 34M parameters.
For ST-only fine-tuning, this model is then fine-tuned on the CoVoST 2 ST dataset. A dropout probability $0.3$ is used on the input embedding and all residual connections in the Transformer decoder to mitigate overfitting.
For multi-modal fine-tuning, this model is fine-tuned on the CoVoST 2 ST dataset simultaneously with the MT dataset described above. Each training batch contains equal numbers of ST and MT examples, with a higher loss weight, $5.0$, on the MT objective.
A lower dropout probability $0.1$ is used because more training data is available.\footnote{ These hyper-parameters were found by optimizing the w2v-BERT speech-only baseline for CoVoST 2 development BLEU.}

\subsection{Speech Classification}
\label{subsec:tasks-sc}
\textbf{Fleurs-LangID:} Fleurs\footnote{Dataset to be released with another publication.} is a speech extension of the FLORES massively multilingual benchmark for MT~\citep{goyal2021flores}. Fleurs contains $2009$ sentences from the FLORES multi-way parallel evaluation set in $102$ languages. We collect read speech corresponding to these sentences, and split these utterances into train-dev-test splits with $1109$ for training (around 1.3 hours of data), $400$ for dev and $500$ for test, per-language. We collected  $2.3$ utterances per sentence on average. We evaluate our pre-trained models on Speech Language Identification (LangID) on this dataset.

\textbf{MINDS-14:} MINDS-14~\cite{gerz2021multilingual} is an intent classification task from spoken data. It covers 14 intents extracted from the e-banking domain, with spoken examples in 14 language varieties. We merge monolingual datasets into a  single dataset, with a $30$-$20$-$50$ train-dev-test split. 

\textbf{Fine-tuning setup:} When fine-tuning our models on speech classification we train the multi-modal and speech encoders. Speech input is fed into the speech encoder, and the outputs from the multi-modal encoder are max-pooled together before feeding into a softmax classifier. Optionally a projection layer is applied before pooling. We tune hyper-parameters on dev performance; tuning batch sizes over $\{16, 32, 64\}$, learning rates over $\{2e-6, 4e-6, 2e-5, 4e-5\}$ and projection over $\{None, model\_dim\}$. For MINDS we tune number of epochs over $\{100, 300\}$ and for Fleurs over $\{5, 10, 20\}$. We pick the run with the best dev performance and evaluate on the test split. For MINDS-14, we report the macro-averaged accuracy over all $14$ languages.

\subsection{Multilingual Speech Recognition}
\label{subsec:tasks-asr}
\textbf{VoxPopuli:} Following~\citet{wang2021voxpopuli}, we evaluate on the $14$ languages with more than $1$-Hr of data from the VoxPopuli ASR task.

\textbf{MLS-10Hr:} We report results on the $10$-Hr training split for the MLS task~\citep{pratap2020mls}.

\textbf{Babel:} Following ~\citet{babu2021xls}, we report results on $5$ languages from the Babel-ASR task.

\textbf{Fine-tuning Setup:} We fine-tune our pre-trained encoders with a $2$-layer LSTM~\cite{hochreiter1997long} as a conformer-transducer model, following~\citet{chung2021w2v}. We use a merged grapheme vocabulary based on the task-specific training set for all ASR fine-tuning experiments. We do not use language-model fusion for any experiments. For VoxPopuli and MLS we report results with multilingual fine-tuning, while we fine-tune separate models per language for Babel. Our finetuning parameters follow~\cite{zhang2020pushing}; for the pre-trained encoder, we use a peak learning rate of $3e-4$ with $5k$ warm-up steps, while for the decoder, a peak learning rate of $1e-3$ and $1.5k$ warm-up steps. All finetuning experiments on ASR use a constant $256$ batch size. In practice, these parameters worked well across several tasks and amounts of data.

\subsection{Text Classification}
\label{subsec:tasks-tc}
\textbf{XNLI:} We also evaluate \mslam{} models on the XNLI sentence-pair classification task~\cite{conneau2018xnli} to understand its strengths and weaknesses on text understanding tasks. We evaluate our models under both the zero-shot and translate-train-all settings~\citep{ruder2021xtreme}, and compare performance against mT5~\citep{Xue2021mT5AM}. 

\textbf{Fine-tuning setup:} We train the multi-modal and text encoders on XNLI. We tune batch sizes over $\{16, 32\}$, learning rates over $\{2e-5, 4e-5\}$, projection over $\{None, model\_dim\}$ and number of epochs over $\{3, 5\}$.

\section{Results}
\label{sec:results}

\subsection{Multilingual Speech Translation}
\label{subsec:results-ast}
\insertcovostshort


\textbf{ST fine-tuning:} Multilingual speech translation results are shown in Table~\ref{tab:covost_xen_short}. Removing all text and paired data from pre-training gives us our speech-only pre-training baseline, \wvbert, which is already very competitive with the state-of-the-art, outperforming \xlsrpb{1}~\cite{babu2021xls}, despite having fewer parameters and not using a pre-trained decoder.
\mtlm{} adds text and a paired TLM objective to pre-training as described by \citet{bapna2021slam}, and actually leads to an average degradation in ST quality, potentially due to interference between the speech and text modalities alongside the additional pressure of massive multilinguality.
Fortunately, \mctc's addition of a CTC component to the TLM objective, as described in Section~\ref{sec:pretraining}, recovers \wvbert\ performance; in fact, it improves slightly, mostly on low-resource languages.
As we show in Section~\ref{sec:analysis}, this CTC component is essential to zero-shot cross-modal behavior.

\textbf{ST + MT joint fine-tuning:} The picture becomes more interesting as we introduce MT (text-to-text) data during fine-tuning in the bottom four lines of Table~\ref{tab:covost_xen_short}. On top of the speech-only \wvbert, adding MT data produces a modest average improvement of $+0.6$ BLEU. However, adding MT data to \mctc, results in a larger improvement of $+1.8$ BLEU, suggesting that exposure to text during pre-training makes the encoder more amenable to using text during fine-tuning. This results in a new state-of-the art for the CoVoST 21$\rightarrow$En task, surpassing the 4$\times$ larger \xlsrpb{2} by $0.3$ BLEU, enabled by large gains on high-resource languages. Increasing the capacity of \mctc{} to $2B$ parameters further improves performance by $2.4$ BLEU.


\subsection{Speech Classification}
\label{subsec:results-sc}
\insertsc

Evaluations on the MINDS-14 and Fleurs-LangID tasks are detailed in Table~\ref{tab:sc}. We find that pre-training jointly with text and paired data with a TLM loss, \mtlm, improves over our speech-only baseline, \wvbert, by $1.3\%$ and $4.6\%$ on MINDS-14 and Fleurs-LangID respectively. The addition of a CTC loss in \mctc{} further improves accuracy by $2.9\%$ on MINDS-14. On Fleurs-LangID, \mctc{} is worse than \mtlm{} by around $2.7\%$, still maintaining an accuracy improvement of $1.9\%$ over our speech-only baseline. Increasing \mctc{} capacity to $2B$ parameters results in further $1.7\%$ improvement over our previous best accuracy on Fleurs, while being $0.3\%$ worse than the $600M$ model on MINDS-14.

\subsection{Multilingual Speech Recognition}
\label{subsec:results-asr}
\insertasr

We present ASR results on VoxPopuli, Babel and MLS-10hrs in Table~\ref{tab:asr}. Our speech-only pre-training baseline, \wvbert~already outperforms \xlsrp ~\cite{babu2021xls} on VoxPopuli and MLS-10hrs as shown in Table~\ref{tab:asr}. Our \mctc{} (0.6B) model slightly outperforms the speech-only baseline on VoxPopuli and slightly lags on MLS-10hrs, but both improve over published results. On Babel, our model is behind \xlsrpb{1}~\cite{babu2021xls}; possibly due to a lack of language model fusion. \mctc{} is very close to \wvbert~and both improve over \mtlm. In conclusion, \mslam{} achieves competitive ASR results without losing speech capacity across a variety of ASR tasks and languages. 
Increasing \mctc{} capacity to $2B$ parameters results in improvements over both, the $600M$ model and our speech-only baseline.

\subsection{Text Classification}
\label{subsec:results-tc}
\inserttcshort
On XNLI, similar to SLAM results on GLUE, we observe decreases in performance compared to mono-modal models due to capacity dilution (see Table~\ref{tab:xnli_short}). In the translate-train-all setting, our \mctcb{0.6} model obtains 70.0\% accuracy on average compared to 79.8\% for an mT5-Base model (0.6B). However, it performs comparably to the smaller mT5-Small model (0.3B) which gets 71.3\%. 

On zero-shot classification, we observe a bigger drop in performance when using multi-modal pre-training compared to the mT5 models. Zero-shot classification being a test-bed for the sharing of multilingual representations, we attribute this to speech interfering with the sharing of text representations across languages. Looking more closely at per-language results, the performance drops in particular for non-European languages, e.g. Thai and Chinese where the model loses around $20\%$ accuracy. Note that the paired data used during pre-training is predominantly from European languages, and the performance of \mctc{} improves significantly over \mtlm{} on this set of languages. We hypothesize that having in-language paired data and alignment losses could be contributing to reduced interference between speech and text for these languages, resulting in more robust representations. This is also supported by the significantly improved performance on non-European languages with the \mctcb{2} model, where the increased capacity might alleviate some of the interference. However, there are other confounding factors in our text pre-training approach compared to standard multilingual text pre-training, including the conformer architecture and fully character-level encoder pre-training, which might be contributing to these findings. We leave the study of multilingual representation alignment in joint speech-text models to future work.

\section{Analysis}
\label{sec:analysis}

\textbf{Do we really need text pre-training or just alignment losses?}
\mctc{} models add two improvements over the speech-only baseline: (i) TLM and CTC alignment losses over paired data, and (ii) Pre-training with large amounts of web-text. This raises the question whether our improvements are arising from (i), (ii) or a combination of the two. To answer this question we train a \mctc{} model on unlabeled speech and paired speech text data, but no unlabeled text. We evaluate this model on CoVoST ST, MINDS-14 and Fleurs-LangID and present results in Table~\ref{tab:ablate}. We find that the performance of the \mctc{} model without text falls somewhere between our speech-only model and \mctc{} on MINDS-14 and Fleurs-LangID, suggesting that the additional text pre-training data is at least partially responsible for the observed improvements on these tasks. On CoVoST-2, \mctc{} without text almost matches the performance of \mctc{} when fine-tuning jointly with text translation data, suggesting that a majority of the improvements in this setting arise from MT data, and the alignment loss might be enough to enable the model to benefit from text supervised data for fine-tuning.
\insertablate

\textbf{Are cross-modal representations really aligned?}
\insertanalysis
\insertCtcExamples

We have seen benefits from adding text to pre-training alongside a CTC loss. The importance of this CTC loss suggests that some amount of cross-modal representation alignment is necessary to take advantage of speech and text data in the same model, but can we construct an experiment to clearly demonstrate this alignment?

Zero-shot performance is one strong indicator for representation alignment. To that end, we use our joint fine-tuning infrastructure to conduct CoVoST 2, 21$\rightarrow$En translation experiments where we fine-tune the \mctc{} model on one modality (speech or text) and evaluate on the CoVoST~2 test set using the the other input modality.

Cross-modal results, alongside the amount of paired data available during pre-training, are shown in Table~\ref{tab:crossmodalanalysis}.
For score calibration, the S$\rightarrow$S column shows a modality-matched scenario of fine tuning on speech and testing on speech, corresponding to the sixth row of Table~\ref{tab:covost_xen_short}.
First, note that zero-shot cross-modal translation is possible: the S$\rightarrow$T column shows that fine-tuning on speech and testing on text results in translation performance above 5 BLEU for 13 of 21 languages. Furthermore, 6 of those 13 languages had no paired data available during pre-training, demonstrating the power of being both multimodal and multilingual.
Most surprisingly of all, Russian (ru) has an excellent zero-shot score of 21.9 BLEU, and it not only has no paired data during pre-training, but also no paired data in its Cyrillic script, yet the \mslam{} model can translate it into English. 

We tested for the same behavior with \wvbert\ and \mtlm{} and found no evidence of zero-shot S$\rightarrow$T transfer.
We also tested the impact of unlabeled text:
average zero-shot BLEU is 9.4 for full \mctc{} but only 6.4 without any unlabeled text during pre-training (not shown). The languages without paired data suffer disproportionately: Swedish (sv) drops from 15.2 to 0.7 BLEU and Russian (ru) drops from 21.9 to 4.2 BLEU.

There is still much work left to be done. 
Russian is somewhat of an outlier in terms of script sensitivity, all other cross-modal success stories are for predominantly European languages in the Latin script. Furthermore, some languages such as Turkish (tr) have paired data available during pre-training, but demonstrate limited zero-shot transfer. Finally, note that this zero-shot transfer does not work in the other direction: the T$\rightarrow$S column clearly shows that a system fine-tuned only on text cannot translate speech.

\insertTranslationExamples
\textbf{Examining zero-shot text translation outputs.}
 While the system's cross-modal capabilities are surprising, there is still a substantial drop in BLEU for zero-shot translation of text: compare the S$\rightarrow$S column to the S$\rightarrow$T column in Table~\ref{tab:crossmodalanalysis}. This reduced performance often manifests as repeated or empty outputs: see Table~\ref{tab:translationExamples} for contrastive zero-shot text translation examples. This is reminiscent of oscillatory hallucinations caused by unexpected inputs~\cite{Lee02018hallucinations}.

\textbf{Visualizing cross-modal alignment with a CTC probe.}

To visualize the information available when text is input to a model fine-tuned only for speech, we create a CTC probe for \mslam{} encodings. Freezing the \mslam{} encoder after pre-training, we  tune only the softmax parameters of a CTC decoder using a 21-language ASR objective on the CoVoST 2 data: speech is input, and the gold character-level transcription is the output. We can then decode the CoVoST~2 test set using either speech or text inputs. If speech is input, the ASR task matches the fine-tuning objective. If text is input, this represents a zero-shot character-level auto-encoding (CAE) task. We measure the success of both tasks using character-error-rate (CER).

Per-language results of this CTC probe are also shown in Table~\ref{tab:crossmodalanalysis} as S$\rightarrow$T CAE. First, it is notable that even with a frozen encoder and a far less powerful decoder, we still see zero-shot transfer from speech to text inputs. In fact, with the exception of Turkish (tr) and Russian (ru), zero-shot CAE performance with less than 20 CER is predictive of zero-shot translation performance greater than 5 BLEU. Russian again is an interesting case, with terrible zero-shot CAE performance, but excellent zero-shot MT performance. However, the real value of such a probe is the ability to inspect the outputs. Table~\ref{tab:analysisExamples} shows randomly selected examples for both, a typical success (French, fr) and one of our more mysterious languages (Russian, ru). For French, most of the content is retained with text input, though some capitalization and punctuation is lost. Interestingly, for Russian, its Cyrillic text input results in a partial transliteration into the Latin script.  This suggests that Russian is mapped into the same encoding space as Latin script languages during pre-training, which helps explain its strong cross-modal and cross-lingual transfer behavior.







\section{Conclusion}
\label{sec:conclusions}
We introduced \mslam{}, a multilingual pretrained model capable of representing speech and text in a shared representation space. \mslam{} is trained on unlabeled speech in $51$ languages with a w2v-BERT objective and character-level text in $101$ languages with a SpanBERT objective. In addition to unlabeled data, we train \mslam{} on small amounts of paired speech-transcript data with a novel TLM+CTC objective to encourage representation sharing across the two modalities. Downstream evaluations on CoVoST 2 Speech Translation, Speech intent classification and Speech LangID demonstrate that \mslam{} improves over equivalent speech-only baselines on speech understanding tasks, while maintaining similar quality on ASR. In addition to fine-tuning with labeled speech data, \mslam{} can also leverage text supervision to improve the quality of end-to-end speech tasks, as demonstrated by our experiments on CoVoST 2 Speech Translation, establishing a new state of the art on this dataset. Increasing the capacity of \mslam{} to $2B$ parameters further improves quality on Speech Translation, Speech Language Identification and multilingual ASR.

On XNLI sentence-pair classification, we observe cross-lingual zero-shot performance equivalent to text-only models half the size of \mslam{} on (European) languages with relatively large amounts of paired data, but severe quality degradation on languages with scarce parallel data. We notice that this degradation can be addressed to some extent by increasing the capacity of the joint speech-text model. 

When fine-tuned on speech translation only, \mslam{} is capable of cross-modal zero-shot text translation, demonstrating strong evidence for representation alignment. Probing a frozen \mslam{} encoder with a CTC head fine-tuned for ASR demonstrates high quality on text reconstruction, providing additional supporting evidence in favour of aligned speech-text representations.

The use of paired data and alignment losses results in quantitative improvements on several speech understanding tasks and reduced degradation on text understanding tasks, highlighting the need for mitigating interference in multilingual multi-modal pre-training.  We hope that this work catalyzes further research towards improving and understanding universal, multi-modal pre-trained models.

\bibliography{mslam}
\bibliographystyle{icml2022}

\newpage
\appendix

\onecolumn

\insertcovost

\insertvp
\insertbabel
\insertmls
\inserttc

\end{document}